\def\year{2022}\relax
\pdfoutput=1
\documentclass[letterpaper]{article} 
\usepackage{aaai22}  
\usepackage{times}  
\usepackage{helvet}  
\usepackage{courier}  
\usepackage[hyphens]{url}  
\usepackage{graphicx} 
\usepackage{natbib}  
\usepackage{caption} 
\DeclareCaptionStyle{ruled}{labelfont=normalfont,labelsep=colon,strut=off} 
\usepackage{algorithm}
\usepackage{algorithmic}

\usepackage{newfloat}
\usepackage{listings}
\floatstyle{ruled}
\newfloat{listing}{tb}{lst}{}
\floatname{listing}{Listing}

\usepackage{amssymb}
\usepackage{comment}
\title{Targets in Reinforcement Learning to solve Stackelberg Security Games}
\author{
    Saptarashmi Bandyopadhyay
    \textsuperscript{\rm 1},
    Chenqi Zhu
    \textsuperscript{\rm 1},
    Philip Daniel
    \textsuperscript{\rm 2},
    Joshua Morrison
    \textsuperscript{\rm 2},
    Ethan Shay
    \textsuperscript{\rm 3},
    John Dickerson
    \textsuperscript{\rm 1}
}

\affiliations{
    \textsuperscript{\rm 1}
    Department of Computer Science, University of Maryland, College Park, Maryland 20742\\
    saptab1@umd.edu, chqzhu@umd.edu, johnd@umd.edu
    
    \textsuperscript{\rm 2}
    Montgomery Blair High School, Silver Spring, Maryland 20901\\
    philip.daniel.322@gmail.com, joshuaevm@gmail.com

    \textsuperscript{\rm 3}
    Wootton High School, Rockville, Maryland 20850\\
    ethanshay888@gmail.com
    
}

\usepackage{bibentry}

\begin{document}

\maketitle

\begin{abstract}
Reinforcement Learning (RL) algorithms have been successfully applied to real world situations like illegal smuggling, poaching, deforestation, climate change, airport security, etc. These scenarios can be framed as Stackelberg security games (SSGs) where defenders and attackers compete to control target resources. The algorithm's competency is assessed by which agent is controlling the targets. This review investigates modeling of SSGs in RL with a focus on possible improvements of target representations in RL algorithms. 
\end{abstract}

\section{Introduction}

Reinforcement Learning (RL) is useful to solve real world problems whose models are inspired by game theory (GT). Stackelberg security games (SSGs) in GT formulate real world agents competing to control targets in domains like protecting infrastructure 
\citep{9668430}, preventing drug smuggling, combating illegal fishing and 
animal poaching, mitigating illegal logging, and other climate change problems \citep{9722864}. 
Successful automated assessment of RL algorithms for SSGs require well-defined and interpretable target (secure resources) representations.
\subsection{Stackelberg Security Games}
Stackelberg security games (SSGs) are two-player games with one defender and one attacker. The defender aims to protect a set of targets $T$ by allocating limited resources to protect up to $b$ targets, where $b\subset T$. The payoff functions for the defender and attacker are $U^d(t)$ and $U^a(t)$ respectively for $t \in T$. $\bold{p}_t$ is the probability of target $t$ to be protected by the defender and $\bold{q}_t(\bold{p},\xi)$ is the probability of target $t$ being attacked by the attacker with feature $\xi$. \citep{1499829}. The two-player SSG has been extended to a coalitional security game \citep{guo2016coalitional} or single leader multiple follower SSG by using a cardinal function of agent utilities \citep{9722864}.



\subsection{Targets for Competency Self-assessment}
Traditionally, the performance of the autonomous agents in MARL can be evaluated using reward signal such as Bellman equation \citep{Uther2010}. \citet{arxiv.2101.05507} proposes a novel technique for testing the robustness of MARL agents where ``unit tests" are designed to gauge the agents' abilities to solve specific tasks. A Strong Stackelberg Equilibrium (SSE), where the attacker breaks ties in the best response set in favor of the defender, is the standard solution concept for SSGs \citep{von2004leadership} due to no guarantees about the existence of a Weak Stackelberg Equilibrium (WSE) \citep{conitzer2006computing}. Even the SSE may not be applicable for security games, as the defender may not induce the attacker to select a preferred action with an infinitesimal change in their policy due to resource assignment constraints. \citet{guo2019inducibility} proposes the inducible Stackelberg equilibrium (ISE), defining an inducible target that serves as the lower bound of the defender's utility guarantees. SSGs involve targets that the attackers attempt to gain control over and the defenders protect from the attackers \citep{kiekintveld2009computing, sinha2018stackelberg, wang2019using}. The control of targets determines the winner, consequently determining the success of policies in autonomous systems that help defenders to win. This motivates future work to develop a solution concept for SSGs with the notion of targets that can be applied to RL. 

\section{Modeling SSGs in Reinforcement Learning}

\subsection{Markov Decision Process}
Markov Decision Processes (MDPs) are described by a tuple of $(S,A,P,R)$ where $S$ is the set of states, $A$ is the set of actions, $P:S \times A \rightarrow \Pi (S)$ is the state-transition function where $P(s'|s,a)$ is the probability of transitioning from state $s$ to $s'$ given action $a$, and $R:S \times A \rightarrow \mathbb{R}$ is the reward function where $R(s,a)$ is the reward of action $a$ in state $s$ \citep{Bel}. 

For a sequential stochastic SSG with one leader (defender) and one follower (attacker), \citep{https://doi.org/10.48550/arxiv.2005.11853} proposes the Markov Perfect Stackelberg Equilibrium (MPSE), which computes a Markov equilibrium in linear time from a Stackelberg equilibrium in double exponential time using sequential decomposition, and applies it to a MDP. A model-free RL algorithm motivated by Expected Sarsa computes the optimal MPSE strategies for the defender and the attacker, where the defender maximizes its returns based on the knowledge that the attacker is strategizing to play the best response to the defender's strategy. 

\subsection{Partially Observable Markov Decision Process}
Partially Observable Markov Decision Processes (POMDPs) extend the MDP's definition to $(S,A,P,R,\Omega,O)$ where $\Omega$ is the set of observations and $O:S \times A \rightarrow \Pi (\Omega)$ is the observation function where $O(o|s',a)$ is the probability of the agent observing $o$ after action $a$ landing in state $s'$ \citep{KAELBLING199899}.

\citet{ALCANTARAJIMENEZ2020106695} models SSGs as POMDPs where agents cannot access the full information regarding the state. \citet{8753973} finds the optimal defender policies with policy gradient in a partially observable cooperative RL framework. \citet{DBLP:journals/corr/abs-2012-10389} uses  Double Deep Q Learning (DDQN) to compute allocation and patrolling strategies with drones for SSGs with partial observability.

\subsection{Dec-Partially Observable Markov Decision Process}
Decentralized Partially Observable Markov Decision Processes (Dec-POMDPs) generalize POMDPs as $([N],S,A_{i \in [N]},P,R,\Omega_{i \in [N]},O_{i \in [N]}))$. Considering an $m$-sized subset of $N$ agents who can only partially observe the states, Dec-POMDPs update the set of actions $A_i$ for agent $i$, the transition probabilities $P(s'|s,a_1,...,a_m)$ with $[1,m] \subset [N]$, the joint reward $R(s,a_1,...,a_m)$, the set of observations $\Omega_i$ for agent $i$, the observation function $O_i:S \times A_i \rightarrow \Pi (\Omega_i)$ with observation probabilities $O(o_1,...,o_m|a_1,...,a_m,s')$ that denotes the set of agents $i=1...m$ observing $(o_1,...,o_m)$ after a tuple of actions $(a_1,...,a_m)$ landing in $s'$. Dec-POMDPs formulate SSGs in multi-agent RL by modeling decentralized control over agents \cite{10.5555/2073946.2073951}. 

\subsection{Partially Observable Stochastic Games}
Partially Observable Stochastic Games (POSGs) generalize from Dec-POMDPs to $([N],S,A_{i \in [N]},P,R_{i \in [N]},\Omega_{i \in [N]},O_{i \in [N]})$.
It defines reward functions for each agent $i$ as $R_i:S \times \prod_{i \in [N]}A_i  \rightarrow \mathbb{R}$. 
POSGs formulate the multi-agent environment by simultaneous stepping where all agents step, observe, and are rewarded together \citep{10.5555/1597148.1597262,https://doi.org/10.48550/arxiv.2009.14471}. 

\section{Target Representation to Assess RL for SSGs}
Targets are important in SSGs, signifying the essential component of security: Attackers try to gain control from defenders while defenders protect targets from attackers. We analyze how targets are modeled in RL literature for SSGs.

\subsection{Targets as a Part of State}
\citet{8753973} defines targets to be a part of the state, while the objectives for the defenders and attackers are to protect and attack the targets respectively given a state. However, targets can be moved from one state to another \citep{wang2019using}, either as a result of agent actions or due to stochasticity. Existing research work applying repeated SSGs with imperfect information in RL \citep{ALCANTARAJIMENEZ2020106695} considers the targets to satisfy Markov and complex physical constraints where the games' strategies evolve around patrolling the targets which are fixed locations on a map. Likewise, the learning algorithms to solve security games consider the targets to be nodes or cells in a grid-world state space where vulnerable resources are located \citep{Klma2016MarkovSG, AAAIW1817362, Wang_Shi_Yu_Wu_Singh_Joppa_Fang_2019,  DBLP:journals/corr/abs-2012-10389}. 
These proposed algorithms do not define targets explicitly, which should be done for the solution of ISE to hold as defined before. 


\subsection{Targets as a Part of Strategy}
\citet{clempner2022learning} takes a different approach to define the set of full strategy profiles as $X \times Y =(x^1,...,x^n,y^{n+1},...,y^m)$. The defenders choose a strategy $x^\delta \in X$ and attackers choose a strategy $y^\alpha \in Y$ after observing $x$, implying targets as a part of the strategy. This approach relies on the definition that the defender strategy is a probability distribution over a set of actions that protect a limited number of targets and the attacker strategy is a probability distribution over a set of actions that attack a limited number of targets. However, targets are not necessarily associated with agent strategies as they are resources in SSGs.

\subsection{Challenge of No Explicit Definition of Targets}
Existing literature in RL for SSGs has not modeled targets separately. Some papers modeling SSG as MDP \citep{https://doi.org/10.48550/arxiv.2005.11853, Klma2016MarkovSG} or POMDP \citep{ALCANTARAJIMENEZ2020106695} lack an explicit definition of targets. \citet{7799111, TREJO201835, CLEMPNER2022104703} model SSGs as controllable Markov chains without considering target resources while adapting strategies against dynamic behavioral models of attackers. Given the significance of targets, not including its explicit definition introduces obscurity in modeling the payoff function which subsequently leads to ambiguity in the RL reward function.

\subsection{Abstracting Targets}
In practice, RL algorithm would choose a policy that maximizes expected rewards over a highly randomized one. If we can model the embedded SSGs with targets set, resources constraints, and utilities, the game can be solved prior to learning policies to implement the winning strategy. Moreover, low-level decision control in MARL setting need not to be confined to GT, SSGs can be solved with better time and sample complexity by abstracting the control problem. 

\section{Future Work}
Current literature in Reinforcement Learning (RL) to solve Stackelberg security games either ignores targets or considers them to be a state or strategy which does not reflect the solution concept of these games. Given the limitations, it may be beneficial to define the set of targets independent of existing parameters to learn better policies in RL. There are RL algorithms adapted for application in Stackelberg security games where the targets are considered separately as an input \citep{kar2016comparing} to strategize for the defender's policies. These algorithms will need to be extended to multi-agent RL for understanding optimal attacker policies as well.

\bibliography{aaai22.bib}

\end{document}